\documentclass[journal]{IEEEtran}
\usepackage{hyperref}
\ifCLASSINFOpdf

    \usepackage[pdftex]{graphicx}

\usepackage{amsmath}
\usepackage{tabulary}
\usepackage{xcolor}
\usepackage{dblfloatfix}
\usepackage{caption}
\usepackage{subcaption}
\usepackage{longtable}
\usepackage{graphicx}
\usepackage{subcaption}
\usepackage{mwe}
\usepackage{multirow}
\usepackage{multicol}
\usepackage{cooltooltips}
\usepackage{cite}
\usepackage[T1]{fontenc}
\usepackage{enumitem}
\usepackage{authblk}
\usepackage{hyperref}

\definecolor{brightpink}{rgb}{1.0, 0.0, 0.5}
\definecolor{lime-green}{rgb}{0.2,0.8,0.2}

 \hypersetup{
     colorlinks=true,
     linkcolor=blue,
     filecolor=blue,
     citecolor = lime-green,      
     urlcolor=brightpink,
     }
\usepackage[export]{adjustbox}
\usepackage{booktabs} 
\usepackage{hhline}
\usepackage{algorithm}
\usepackage[noend]{algpseudocode}

\usepackage{etoolbox}
\newbool{EARTHPRO}
\boolfalse{EARTHPRO}

\newcommand{\comment}[1]{}
\begin{document}

\newcommand{\mytitle}{SmartScan: An AI-based Interactive Framework for Automated Region Extraction from Satellite Images}
\title{\mytitle}

\author[12*]{Savinay Nagendra\thanks{*~Work completed during internship at Schlumberger-Doll Research, Cambridge, MA. 02319 during May-August 2023.}}
\author[2]{Kashif Rashid}

\affil[1]{Pennsylvania State University, University Park, PA 16802}
\affil[2]{Schlumberger-Doll Research, Cambridge, MA 02139}

\maketitle
\begin{abstract}

The deployment of a continuous monitoring system for methane emission sources on a client facility entails establishing the optimal number and location of fixed point sensors. The planning process, however, can be labor intensive as it takes considerable effort to setup a site and run multiple iterations to fully capture client restrictions. In addition, this process is particularly time-consuming when many sites are to be evaluated, considerably hindering scalability.

Motivated by this, we introduce SmartScan, an AI framework that automates the extraction of pertinent data-sets necessary for optimal sensor placement design. Thus, the subspaces of interest are extracted from a satellite image of the site with an interactive tool that helps to quickly create facility-specific and problem-dependent constraint sets. 

SmartScan employs the Segment Anything Model (SAM), a prompt-based transformer model for zero-shot segmentation for extracting sub-spaces (regions of interest) from any satellite image without need for explicit training. SmartScan has two modes of operation: (1) In the Data Curation Mode a satellite image is processed to extract high-quality sub-spaces. For this, SmartScan has a unique interactive prompting design to rapidly gather user-prompts for SAM. The extracted sub-spaces are utilized in downstream algorithms. (2) In the Autonomous Mode, user-prompts gathered in the data curation mode are used as ground-truth to train a novel deep learning network. The trained network is deployed to replace user-prompting. Here, the end-to-end subspace extraction process is completely autonomous. Subsequently, the interactive visualization and annotation tool is used for (1) to quality check and correct errors of the AI framework (e.g. to remove false positives that will affect the accuracy of downstream algorithms, and (2) to generate additional facility-specific constraint sets as required. SmartScan is streamlined for producing high-quality sub-space extraction with high throughput and minimal human supervision (quality check) with its novel end-to-end design and AI-based prompting mechanism, thus increasing scalability and efficiency of downstream algorithms. Notably, the design of SmartScan makes it suited for extracting regions of interest from any ultra high-resolution satellite imagery, making it domain agnostic.

\end{abstract}

\IEEEpeerreviewmaketitle

\section{Introduction}\label{Sec:1}
The oil and gas industry is facing an increasing demand to monitor its assets for methane leaks as part of efforts to reduce greenhouse gas emissions \cite{collins2018increased}. Methane is a potent greenhouse pollutant, with a global warming potential 84 times greater than that of carbon dioxide \cite{collins2018increased}. Approximately 20\% of annual anthropogenic emissions can be attributed to the oil industry \cite{scarpelli2020global, rashid2023subspaceconstrained}. These emissions fall into two main categories: intentional venting or unintentional fugitive leaks. Intentional venting occurs as a result of operational activities where methane is knowingly released into the atmosphere (e.g., resulting from the use of pneumatic natural gas valves or direct venting)\cite{soltanieh2016review}. While such leaks are undesirable, they can be addressed with revised work practices and the use of equipment that eliminates emissions. In contrast, fugitive leaks result from malfunctioning equipment such as wellheads, separators, compressors, and pipelines \cite{soltanieh2016review}. Recent research indicates that a small number of these leaks are responsible for a significant portion of total emissions \cite{cusworth2021intermittency, zavala2017super, brandt2016methane}. Hence, there is a pressing need to rapidly identify and repair sources of methane pollution, and especially those identified as super-emitters\cite{rashid2023subspaceconstrained}.

\begin{figure}[htpb]
    \centering
    \includegraphics[width=\linewidth]{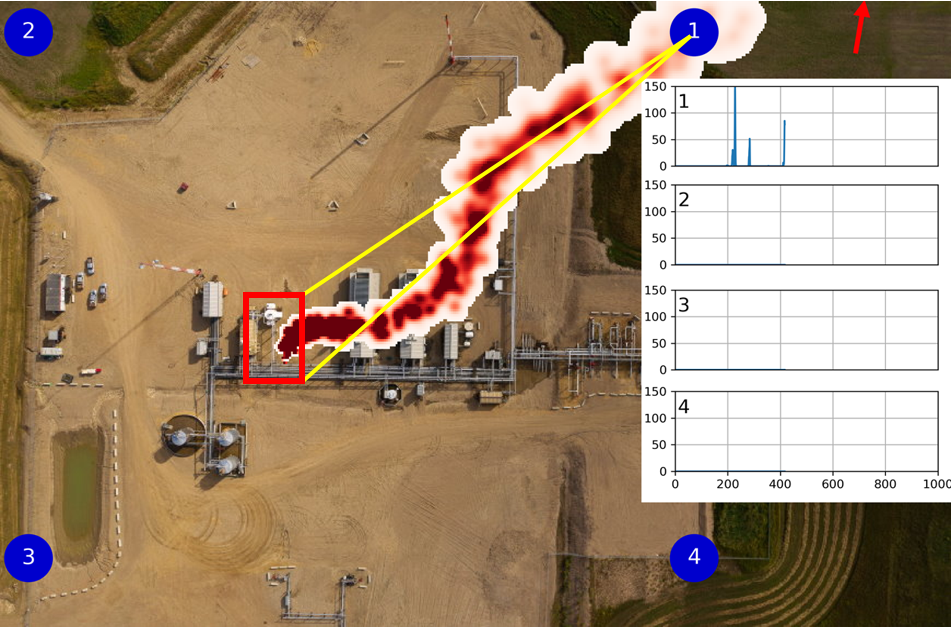}
    \caption{\textbf{Example of a continuous methane leak monitoring system.} An optimal number of sensors have been deployed at the facility according to the generated optimal placement design. Further, targeted source-inversion is used to determine the subspace inside which a leak occurs.}
    \label{fig:cont-system}
\end{figure}

\begin{figure*}[htpb]
    \centering
    \includegraphics[width=\linewidth]{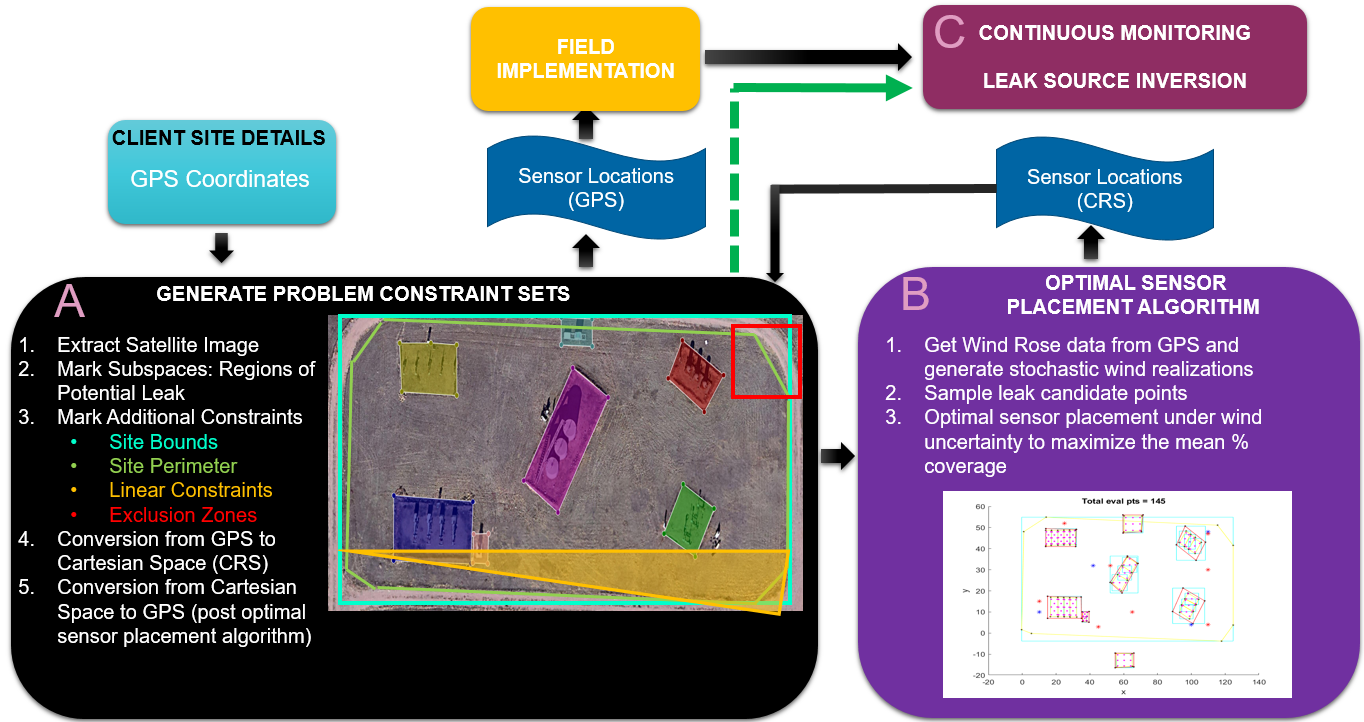}
    \caption{\textbf{Continuous real-time monitoring system workflow.} Client-provided GPS coordinates are given to module \textit{\textbf{A}}, where facility-specific constraint sets are generated. These are given to module \textit{\textbf{B}} to generate an optimal sensor placement design. The proposed sensor placement design is converted to GPS coordinates, and deployed at the facility. Finally, targeted source-inversion happens in module \textit{\textbf{C}} for identifying the subspace in which a leak occurs with continuous monitoring.}
    \label{fig:workflow}
\end{figure*}

Continuous real-time monitoring has been identified as the most desirable strategy for detecting sources of methane leak, and is an absolute necessity for climate emissions control \cite{rashid2023subspaceconstrained}. Such a continuous real-time monitoring system \cite{rashid2023subspaceconstrained} developed at Schlumberger-Doll Research (SDR) utilizes permanently installed low cost metal-oxide methane sensors at a facility that can continuously monitor and identify leaks, with source localization and quantification methods to aid expedient remediation. The deployment of such a continuous monitoring system, as shown in Figure \ref{fig:cont-system}, entails (1) Defining subspaces (the regions containing potential sources of methane leak), (2) Establishing the optimal number and location of the fixed point sensors, and (3) Targeted source-inversion to find the subspace inside which a leak occurs during active monitoring. 

\begin{figure}[htpb]
    \centering
    \includegraphics[width=\linewidth]{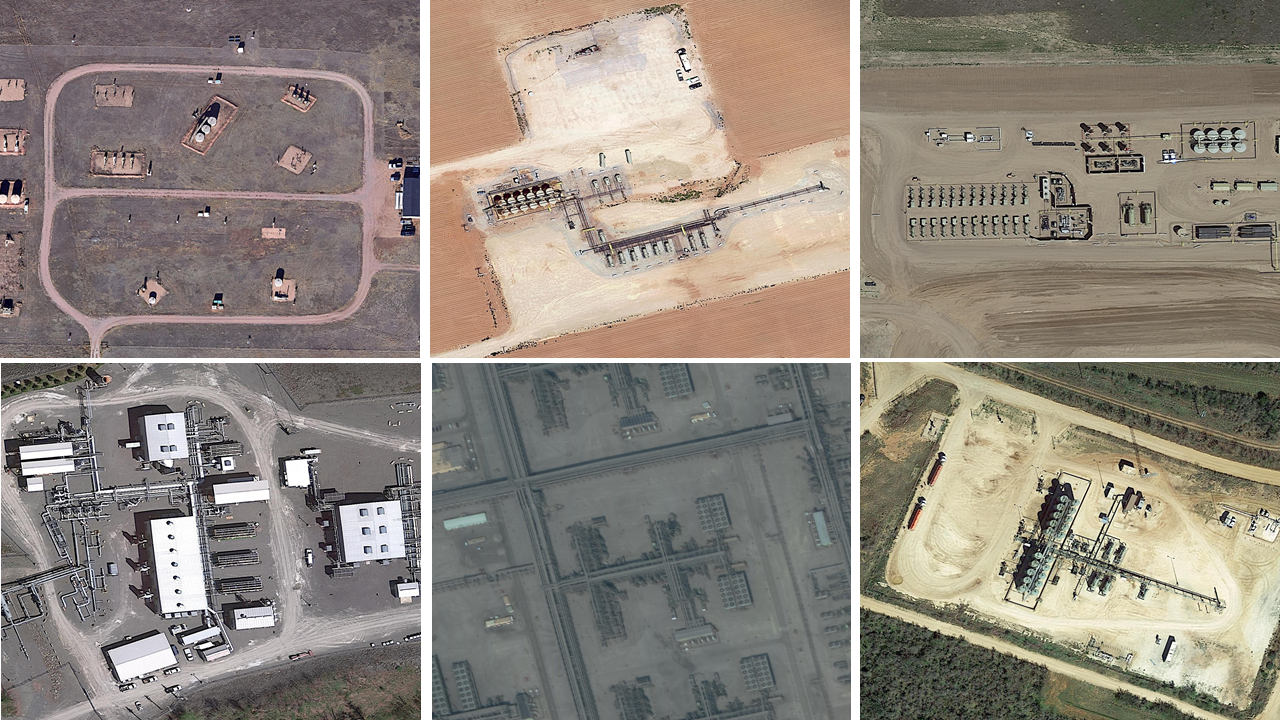}
    \caption{\textbf{Satellite images of different client facilities.} It can be observed that each facility is visually dissimilar from one-another, which makes the process of manually defining subspaces labor intensive.}
    \label{fig:sat-imgs}
\end{figure}

The entire workflow of a continuous real-time monitoring system is shown in Figure \ref{fig:workflow}. The client provides GPS coordinates of the facility which is given to module \textbf{\textit{A}} as an input. In this module, a high resolution satellite image is extracted from the GPS coordinates. Subspaces (regions of potential leak) are drawn manually. Further, additional constraints such as site bounds, site perimeter, linear constraints and exclusion zones are also marked that collectively define feasible and infeasible regions for sensor placement. Subspaces, together with these constraints make a facility-specific problem constraint set. These are converted from pixel coordinate system to Cartesian coordinate system (CRS) and given to module \textbf{\textit{B}} in the form of JSON files. Module \textbf{\textit{B}} is for optimal sensor placement. First, wind rose \cite{roubeyrie2018windrose} data is extracted given client-provided GPS coordinates to generate stochastic wind realizations. Next, leak points are sampled from the provided subspaces. Each leak point represents a potential leak source within the subspace that becomes a data point in the optimization. Next, a sensor design is created (number of sensors and their locations). Each leak point is activated and tested whether it is detectable by the set of the sensors in the current design. This is repeated for all candidate leak points to give the mean percentage coverage under wind uncertainty. The design is modified, and the process continues until the mean percentage coverage is maximized. The optimized sensor placement design is returned to \textbf{\textit{A}} which converts the design back to GPS. This final design is conveyed for field implementation of the fixed point sensors at the facility. When the deployed sensors are activated, they start continuously monitoring the facility for leaks. Module \textbf{\textit{C}} hosts the targeted source-inversion algorithm that finds the subspace in which a leak occurs (details  can be found in \cite{rashid2023subspaceconstrained}).

Well-defined subspaces are pertinent to the performance of both the optimal sensor placement and source inversion algorithms (together referred to as downstream algorithms). However, the planning process that entails manually defining subspaces and facility-specific constraint sets is labor intensive as it takes considerable effort to setup a site and run multiple iterations to fully capture client restrictions. This is particularly time-consuming when many facilities are to be evaluated and each is very different as shown in Figure \ref{fig:sat-imgs}. This bottleneck significantly hinders scalability.

In this work, we introduce the development of \textbf{SmartScan}, an AI-based framework\cite{nagendra2017comparison, nagendra2022constructing, nagendra2015video, funk2018learning, nagendra2024patchrefinenet, liu2021new, pei2021utilizing, pei2020cloud, nagendra2020efficient, zhu2022rapid, nagendra2022threshnet, nagendra2023estimating, nagendra2024samic, nagendra2024emotion, nagendra2024thermal, nagendra2025towards, kifersupplementary, cpicacci} that automates the extraction of pertinent datasets necessary for optimal sensor placement design. Thus, subspaces of interest are extracted from a satellite image of the facility with an interactive tool that helps to quickly create facility-specific and problem-dependent constraint sets. SmartScan employs the Segment Anything Model (SAM) \cite{kirillov2023segment}, a prompt-based transformer \cite{dosovitskiy2020transformers} model for zero-shot segmentation \cite{bucher2019zero, nagendra2022constructing, funk2018learning, nagendra2022threshnet} for extracting subspaces from any satellite image without need for explicit training. SmartScan has two modes of operation: (1) \textbf{Data Curation Mode}: A satellite image is processed to extract high-quality subspaces. For this, SmartScan has a unique interactive prompting design for rapidly providing user-prompts to SAM. Extracted sub-spaces are utilized further in downstream algorithms. (2) \textbf{Autonomous Mode}: User-prompts provided in the data curation mode are used as ground-truth to train a novel deep learning network. The trained network is deployed to replace user-prompting. Here, end-to-end subspace extraction process is completely autonomous. Subsequently, the interactive visualization and annotation tool is used for (1) quality check to correct errors of the AI framework (if any) which could affect the accuracy of downstream algorithms, and (2) generating facility-specific constraint sets. SmartScan is streamlined for producing high-quality sub-space extraction with high throughput and minimal human supervision (quality check) with its novel end-to-end design and AI-based prompting mechanism, thus increasing scalability and efficiency of downstream algorithms. Further, the design of SmartScan makes it suited for extracting regions of interest from any ultra high-resolution satellite imagery, making it domain agnostic.

The key contributions of this work are:
\begin{itemize}
    \item The development of SmartScan; an interactive AI-based framework for automated, few-shot and domain-agnostic segmentation of satellite imagery.
    \item The Data Curation Mode of SmartScan is equipped with a novel, interactive prompting module to rapidly generate user-prompts for extracting high-quality segmentation maps from SAM.
    \item The Autonomous Mode of SmartScan is equipped with a novel deep learning based prompt generator module, trained on user-prompts generated in the Data Curation Mode, that can replace user-prompting, enabling end-to-end high-quality segmentation. 
    \item The Quality Check module is equipped with an interactive annotation and visualization tool for generating high-quality subspaces.
    \item We demonstrate that the proof-of-concept novel autonomous prompting module paired with SAM is a powerful scheme for achieving high-quality few-shot segmentation with high generalizability. This scheme is light-weight, easily trainable with few data points, and memory-efficient, while still being able to exceed or have on-par performance with supervised segmentation models.
\end{itemize}

\section{SmartScan}\label{Sec:3}
In this section, we present the SmartScan workflow.

\subsection{Segment Anything Model (SAM)} \label{sec:sam}
\begin{figure}[htpb]
    \centering
    \includegraphics[width=\linewidth]{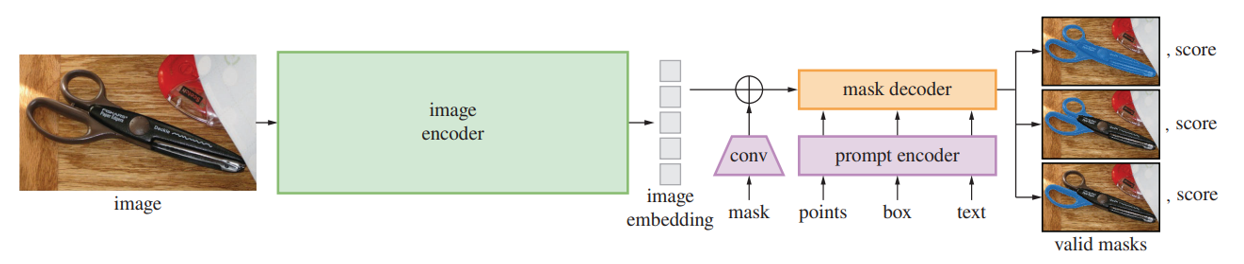}
    \caption{\textbf{Segment Anything Mode (SAM) overview.} A heavy-weight image encoder outputs an image embedding that can be efficiently queried by a variety of input prompts to produce object masks at amortized real-time speed.}
    \label{fig:sam}
\end{figure}

Segment Anything project \cite{kirillov2023segment} is a new task, model, and dataset for image segmentation. Using an efficient model in a data collection loop, Meta AI built the largest segmentation dataset to date (by far), with over 1 billion masks on 11M licensed and privacy respecting images. The pipeline is shown in Figure \ref{fig:sam}. The encoder is an encoder of ViT transformer for extraction. The model is designed and trained to be prompted, so it can transfer zero-shot to new image distributions and tasks. Meta AI evaluated its capabilities on numerous tasks and found that its zero-shot performance was impressive – often competitive with or even superior to prior fully supervised results. Meta AI released the Segment Anything Model (SAM) and corresponding dataset (SA-1B) of 1B masks and 11M images (around April 2023) to foster research into foundation models for computer vision.

\begin{figure*}[htpb]
    \centering
    \includegraphics[width=\linewidth]{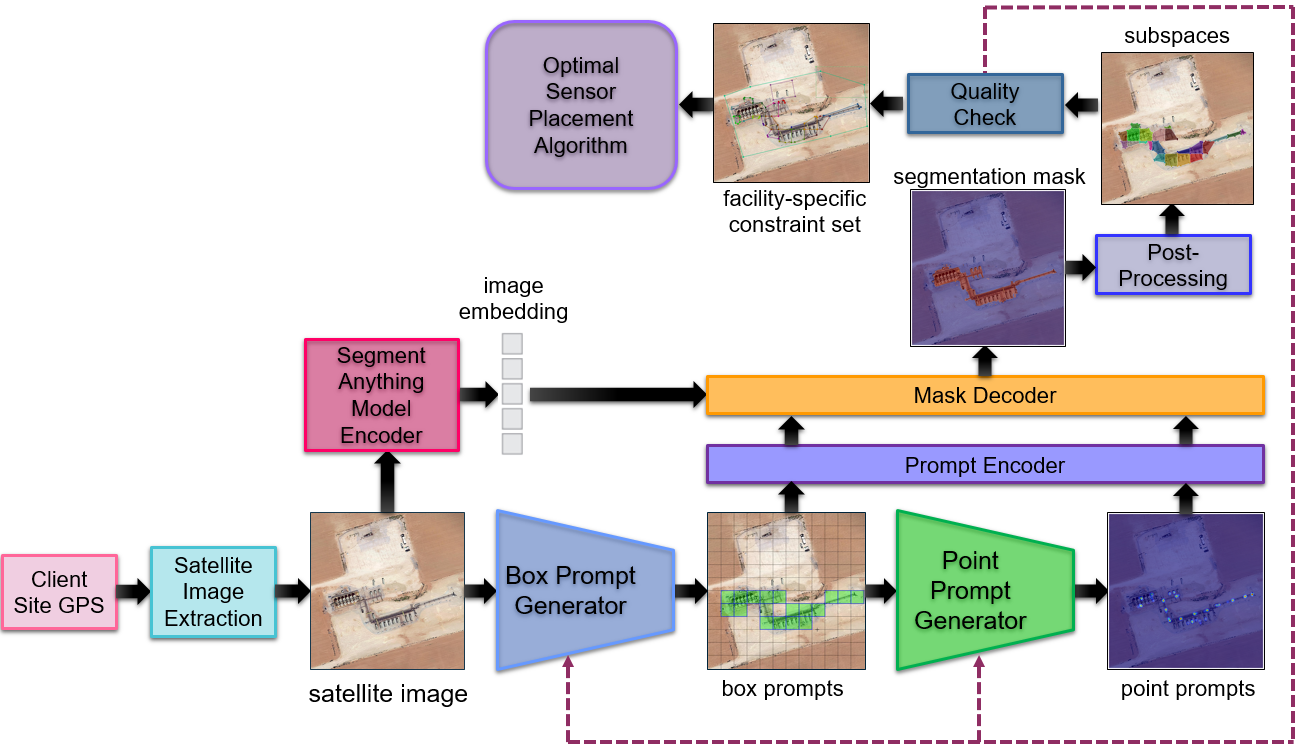}
    \caption{\textbf{SmartScan Back-End Pipeline.} SmartScan is an end-to-end, interactive tool for semi-automated salient region extraction from satellite images. Our tool provides a full stack solution for accurately extracting convex polygons around salient regions of interest, given the GPS coordinates of the site. At its core, SmartScan uses the Segment Anything Model (SAM) for prompt-based segmentation of regions of interest from a satellite image. The generated convex polygon set is used for optimal sensor placement design.}
    \label{fig:backend}
\end{figure*}      

\subsubsection{Image Encoder}
Motivated by scalability and powerful pre-training methods, Segment Anything Model (SAM) uses an MAE \cite{he2022masked} pre-trained Vision Transformer(ViT) \cite{dosovitskiy2020image} minimally adapted to process high resolution inputs \cite{li2022exploring}. The image encoder runs once per image and can be applied prior to prompting the model.

\subsubsection{Prompt Encoder}
SAM uses two sets of prompts: sparse (given as points, boxes or text) and dense (as masks). Points and boxes are represented by positional encodings \cite{tancik2020fourier} summed with learned embeddings for each prompt type and free-form text with an off-the-shelf text encoder from CLIP \cite{radford2021learning}. Dense prompts (i.e., masks) are embedded using convolutions and summed element-wise with the image embedding. For this project, we are not considering text and dense (mask) prompts.

\subsubsection{Mask Decoder}
The mask decoder efficiently maps the image embedding, prompt embeddings, and an output token to a mask. This design, inspired by \cite{carion2020end, cheng2021per}, employs a modification of a Transformer decoder block \cite{vaswani2017attention} followed by a dynamic mask prediction head. The modified decoder block uses prompt self-attention and cross-attention in two directions (prompt-to-image embedding and vice-versa) to update all embeddings. After running two blocks, the image embedding is up-sampled and an MLP maps the output token to a dynamic linear classifier, which then computes the mask foreground probability at each image location.

\subsection{SmartScan: Back-End}
In this section, we discuss the components of the SmartScan pipeline, shown in Figure \ref{fig:backend}, and their functionalities. We will be discussing the Autonomous Mode of SmartScan, which is the current mode of implementation.

\subsubsection{Satellite Image Extraction}
The client provides the GPS \cite{el2002introduction} coordinates of the facility/site where the continuous real-time methane monitoring system must be set up. With this location as the center, bottom-left and top-right GPS coordinates are calculated at a particular zoom-level in Google Maps \cite{googleClientLibraries}, which collectively provides the extent of the site in meters. 
\begin{figure}
    \centering
    \includegraphics[width=\linewidth]{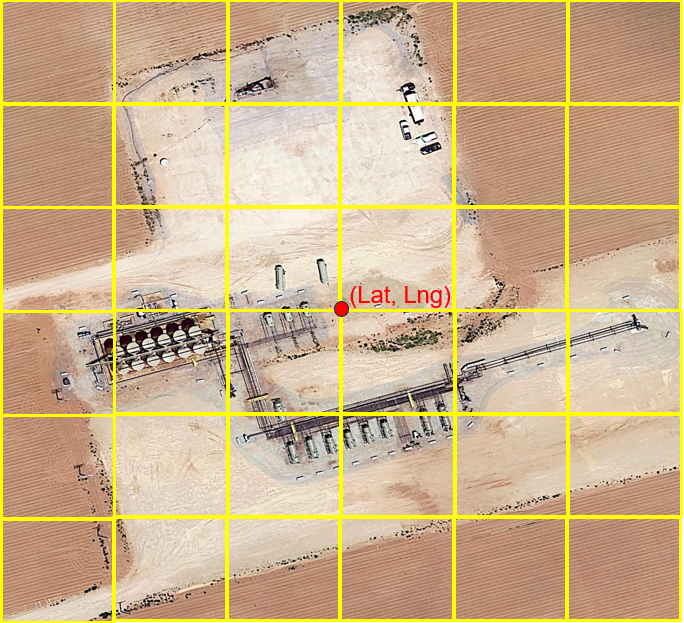}
    \caption{\textbf{Satellite Image Extraction}. An example of ultra-high-resolution satellite image extracted from Google Maps using client-provided GPS coordinates.}
    \label{fig:sat-extract}
\end{figure} The site bounds are converted to pixel domain, where $36$ instances of $512 \times 512 \times 3$ RGB images are stitched together to make one ultra-high-resolution satellite image of size $3072 \times 3072 \times 3$ as shown in Figure \ref{fig:sat-extract}. This image is used to extract regions (subspaces) of potential methane leak sources within the site, and subsequently, to mark additional data necessary for planning purposes (e.g., exclusion zones, linear restrictions, site bounds and the imposed perimeter).

\subsubsection{Prompt Generation}
In this section, we will first discuss how user prompts are collected in the Data Curation Mode of SmartScan. We will then discuss how this collected data from user-prompting is used to develop an autonomous prompting system, which will be used in the Autonomous Mode of SmartScan.\\

\noindent\textbf{Manual Prompting System. } 
After testing the different modes of SAM, as discussed in Section \ref{sec:sam}, we note that both box and point prompts are required for high-quality segmentation outputs, where higher quality segmentation necessarily implies better performance of downstream algorithms.

\begin{figure}[htpb]
    \centering
    \includegraphics[width=\linewidth]{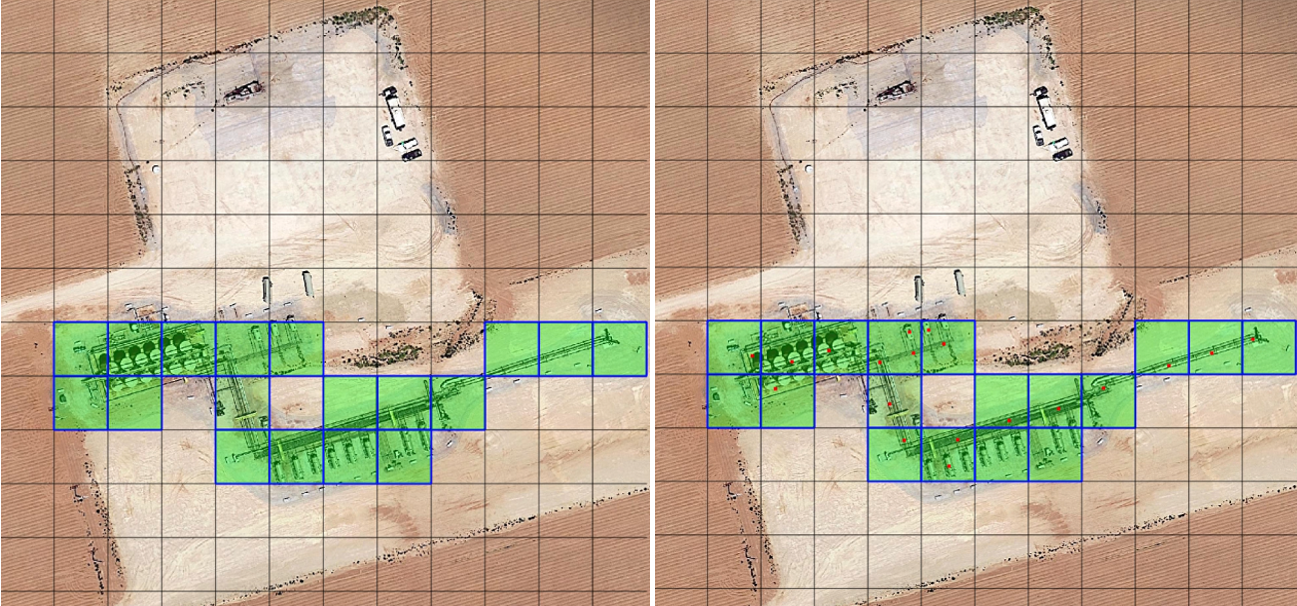}
    \caption{\textbf{Manual Prompting System. } The image on the left shows the user-provided box prompts, marked as green boxes. The image on the right shows one or more user-provided point prompts in every box prompt, marked as red points.}
    \label{fig:manual-prompt}
\end{figure}

We note that SAM is designed to take only one box, and multiple point prompts per image as inputs. However, given the high resolution ($\approx \geq 9MB$) of satellite images, two key challenges are encountered: (1) Giving the entire image to SAM for processing will be significantly memory inefficient, and will fail to work on GPUs with less than 16GB memory capacity. (2) Further, giving a single box prompt for the entire image will result in poor segmentation quality. To alleviate both these challenges, we designed our box prompt system to be similar to CAPTCHA (typically used for web security) as shown in Figure \ref{fig:manual-prompt}. First, the satellite image is presented to the user with $256\times256$ grids. The user clicks on boxes that cover the foreground object (site equipment) and saves the box prompts as a JSON file. Next, the marked box prompts of interest are loaded in another interface where the user marks one or more points in every box prompt. These are also saved as a JSON file. Each $256 \times 256$ grid of the image that was marked by the user now becomes an input to SAM with its corresponding point prompts. This way, all the grids can be processed in parallel on a GPU with increased memory efficiency to obtain high segmentation quality. 
\begin{figure}[htpb]
    \centering
    \includegraphics[width=\linewidth]{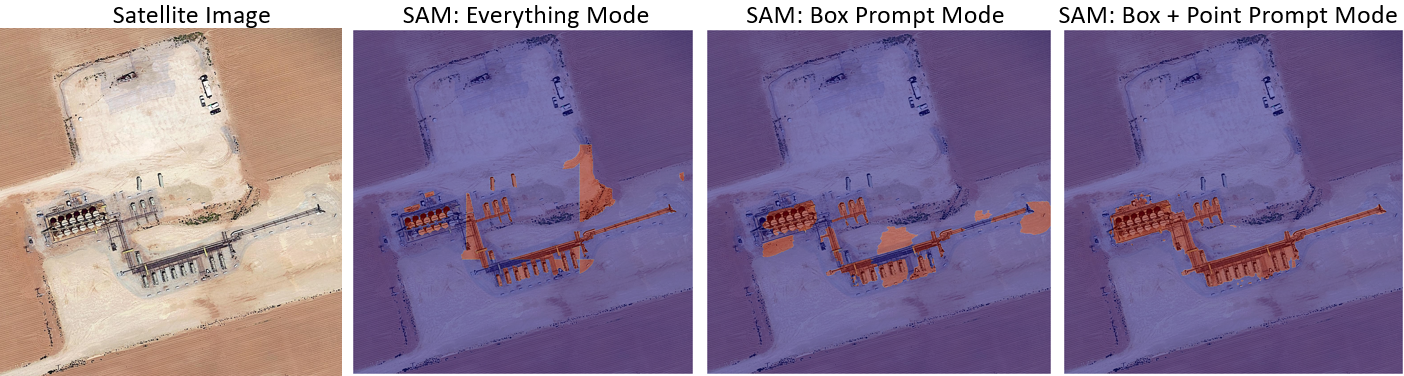}
    \caption{\textbf{Qualitative segmentation results for different modes of prompting SAM.} It can be observed that our manual box + point promoting system achieves the best segmentation quality.}
    \label{fig:prompt-results}
\end{figure}
A qualitative comparison of segmentation outputs from SAM with different modes of prompting is shown in Figure \ref{fig:prompt-results}. It can be observed that our manual box + point promoting system achieves the best segmentation quality. This prompting system is used in the Data Curation Mode of SmartScan.\\

\noindent\textbf{Autonomous Prompting System. } The prompt JSON files stored in the Data Curation Mode contain valuable information regarding the most spatially-important regions of the satellite image. This information can be leveraged to train a light-weight neural network model to automatically pick spatially important regions from satellite images, thereby, learning to prompt SAM, instead of providing manual prompts. This idea is used in the Autonomous Prompting System. 
\begin{itemize}
    \item \textbf{Box Prompt Generator}: The Box Prompt Generator is a simple convolutional binary classifier. It takes as input each $256 \times 256$ grid of a satellite image and assigns a binary class to it: $0$ being \textit{not-of-interest} and $1$ being \textit{of-interest}. With this, all the grids containing regions of interest can be extracted for a given satellite image. Being a binary classifier, the model is light-weight, and can be quickly trained to convergence. The model also has high generalizability and can be trained with less data to provide very high accuracy, making it few-shot.
    
    \begin{figure}[htpb]
        \centering
        \includegraphics[width=0.8\linewidth]{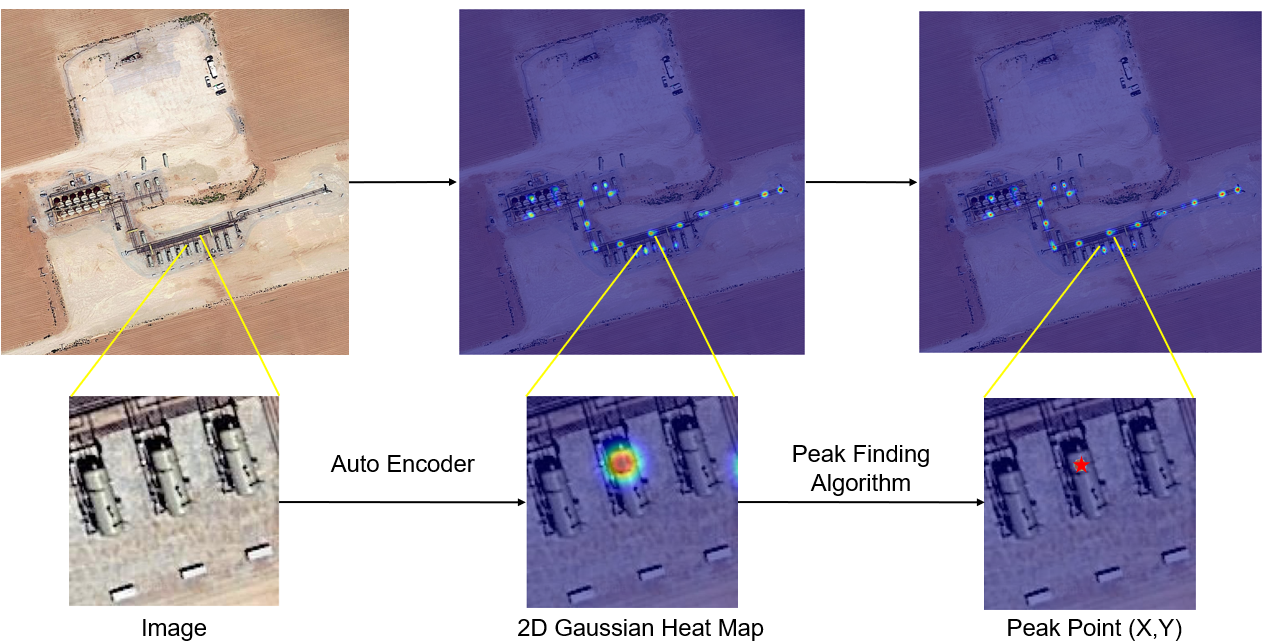}
        \caption{\textbf{Point Prompt Generator. } The Point Prompt Generator is an Autoencoder that takes as input an image grid and outputs a 2D Gaussian heat map that represents the distribution of point prompts in the grid. Finally, a customized peak finding algorithm is used to find the peak of the predicted Gaussians to get the point prompts.}
        \label{fig:point-prompt}
    \end{figure}

    \item \textbf{Point Prompt Generator}: 
    A point prompt is a pixel coordinate $(x,y)$ in an image. Learning a single point using a neural network makes the training very rigid as the model is constrained to predict this single point accurately. A user would mark a point prompt at different pixel coordinates (but in a neighborhood) during multiple iterations of manual prompting. However, we empirically observe that the output of SAM is less sensitive to the location of the point prompt as long as it is within the object of interest. With this motivation, we create a 2D Gaussian heat map of the satellite image as target for training a neural network for learning point prompts. This heat map represents the distribution of all the point prompts, with the center of each Gaussian being the most desired point prompt. We choose the standard deviation of the 2D Gaussians such that segmentation outputs from SAM is not significantly affected when the point is chosen to be within 1 standard deviation away from the center. Our Point Prompt Generator is shown in Figure \ref{fig:point-prompt}. We design the neural network to be an Autoencoder. where the input is an image grid, and output is the corresponding 2D Gaussian heat map. Finally, we employ a customized peak-finding algorithm to find the peak of all the predicted Gaussians to get the point prompts. Again, it has to be noted that this model is light-weight and easily trainable to convergence. The model is also highly generalizable and few-shot as well. 
\end{itemize}

\begin{figure}[htpb]
    \centering
    \includegraphics[width=\linewidth]{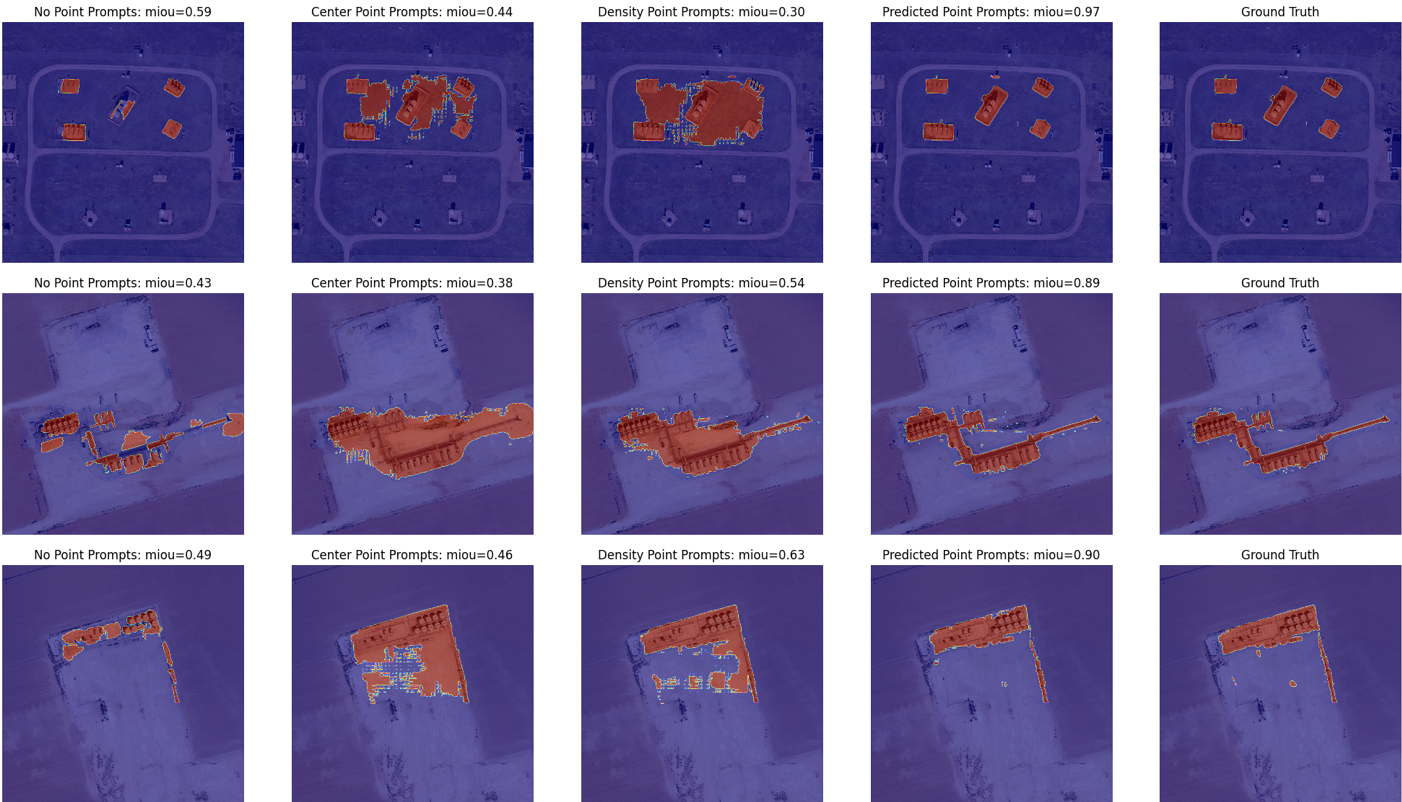}
    \caption{\textbf{Qualitative comparative results of our Autonomous Prompting System. } Segmentation Maps are shown for different prompting techniques. It can be observed that the Autonomous Prompting System performs the best, even on unseen satellite images. Scheme by column: 1) Entire image. 2) Center points. 3) Density-based. 4) Proposed scheme. 5) Ground truth. The error metric given above is the mean intersection over union with respect to the ground truth. The rows represent three different test cases. }
    \label{fig:qual-prompt}
\end{figure}

SAM, combined with our Autonomous Prompting System, can produce high quality segmentation maps, which are on-par or even better than task-specific semantic segmentation models trained from scratch. Qualitative comparative results showing the performance of our Autonomous Prompting system is shown in Figure \ref{fig:qual-prompt}. Two baseline prompt-predicting methods (center points or density-based) and one no-prompt (everything mode) are used for comparison. It can be observed that our Autonomous Prompting System (column 4) gives the best results (both qualitative and quantitative) as shown in Figure \ref{fig:qual-prompt}, with $\approx \geq 90\%$ mean Intersection over Union, even on unseen cases.

\vspace{3mm}
\subsubsection{Post-Processing for Subspace Extraction}
After we get a binary segmentation map from SAM, the next step is to divide this map into subspaces, each made up of tightly-bound convex hulls. The post-processing involves the following steps:
\begin{itemize}
    \item Apply Conditional Random Fields (CRF) \cite{sutton2012introduction} to the segmentation map to get rid of spurious islands of pixels.
    \item Extract contours of connected components using OpenCV.
    \item Filter any left-out spurious islands of pixels by area of extracted contour.
    \item Extract convex hulls from contours of connected components using Sklansky algorithm.
    \item Make convex hulls tight around each object by reducing dead-space (background pixels).
    \item Simplify the convex hulls using the Ramer–Douglas–Peucker \cite{saalfeld1999topologically} algorithm.
\end{itemize}

\begin{figure}[htpb]
    \centering
    \includegraphics[width=\linewidth]{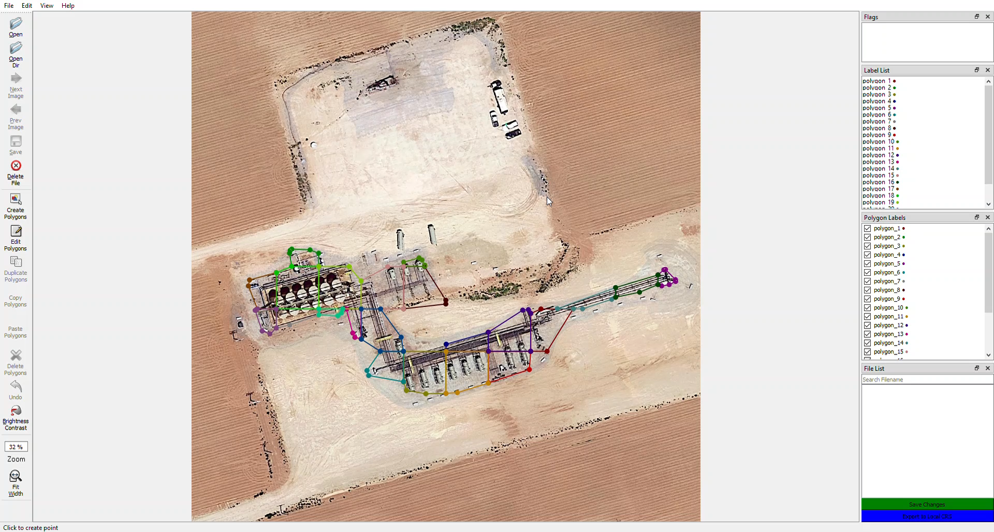}
    \caption{\textbf{Quality Check Tool. } A screenshot of our interactive visualization and annotation tool with multiple functionalities, that enable quick quality check and easy addition of facility-specific constraint sets.}
    \label{fig:qc}
\end{figure}

\vspace{3mm}
\subsubsection{Quality Check}
The Quality Check step is necessary to correct the mistakes (if any) made by the AI models, and to add facility-specific constraint sets as needed for the placement problem of interest. For this, we have designed an interactive visualization and annotation tool as shown in Figure \ref{fig:qc} with the following functionalities:
\begin{itemize}
    \item Create Polygon: Different shapes such as rectangles, circles, ellipses or straight lines can be easily drawn by mouse-drag. Any complex shape can be drawn.
    \item Delete Polygon: False-positive polygons can be deleted with a simple button-click, along with any undesired ones. 
    \item Fragment Polygons: A convex polygon can be divided into four smaller convex polygons to get a tighter bound around regions of interest.
    \item Merge Polygons: Multiple smaller neighboring convex polygons can be merged into a single convex polygon.
    \item Edit Polygons: Vertices can be adjusted with mouse-drag to re-orient/reshape polygons, vertices can be removed to make the polygons simpler, and vertices can be added to form complex tighter shapes.
    \item Defining Elements: 
    The site bounds, site perimeter, subspaces and exclusion zones are marked by polygons.
    A linear constraint can be defined by a triangle, where the first two points indicate the cut and the third point identifies the infeasible half-space. 
    These elements are exported in associated JSON files (as site, subspaces, zones and linear constraints). 
\end{itemize}

\subsection{SmartScan: Front-End}
\begin{figure}[htpb]
    \centering
    \includegraphics[width=\linewidth]{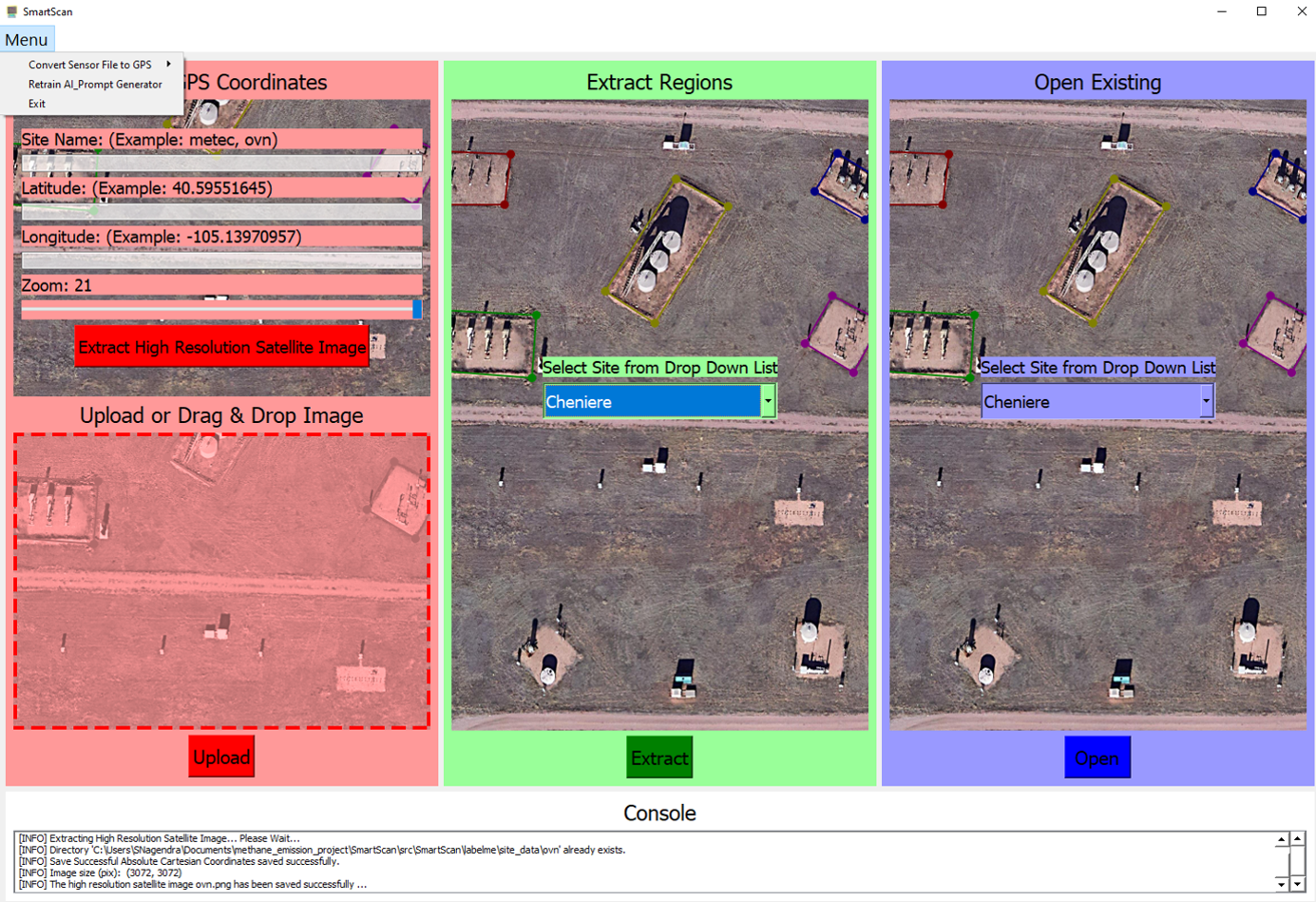}
    \caption{\textbf{SmartScan Front-End.} This image shows the front end design of the SmartScan app. The app has an \textcolor{red}{input module} for providing GPS coordinates, an \textcolor{green}{AI module} for prompt generation and subspace extraction, and a \textcolor{violet}{visualization and annotation module} for Quality Check. The console displays information about background processes executing in real-time.}
    \label{fig:frontend}
\end{figure}

SmartScan is an interactive application for automated subspace extraction from satellite images. The app has been made available on both Linux and Windows operating systems. The master-screen of SmartScan is shown in Figure \ref{fig:frontend}. 

The app has an input module for entering site name, and the client-provided GPS coordinates. This module has three parameters: Latitude, Longitude and Zoom depth. An example of how to enter the coordinates is shown above the field-entry boxes. On the \textit{Extract} button click, a back-end process is called to extract a high-resolution satellite image from Google maps, for the given latitude, longitude and zoom parameters. The zoom will control the spatial extent of the satellite image. If the satellite image extraction fails, a message is provided in the console indicating that the user should change the zoom level and try again. The module has been tested to work for zoom levels of 19 to 21. After the satellite image is extracted, a folder with the same name as the site name will be created. This folder will host all the meta data for that site. The drop down lists will be populated with the site names in both subspace extraction and visualization modules. After the satellite image has been extracted, a user-prompt module will open to provide user-prompts (box and point). The user must provide these prompts and click \textit{save and export} button to save the user-prompts in the folder. This is followed by the user selecting the site and clicking on \textit{Extract}. This will invoke the Segment Anything Model to extract the mask. The mask is post-processed to generate convex polygons. Finally, the user can select from the list of cases in which the subspaces have been extracted and click on \textit{Open} to begin the quality check using the annotation tool. Finally, after the quality check is completed, the user will click the \textit{Save and Export} button to export the JSON files to the designated site folder.

\subsection{Qualitative Results}
\begin{figure}[htpb]
    \centering
    \includegraphics[width=\linewidth]{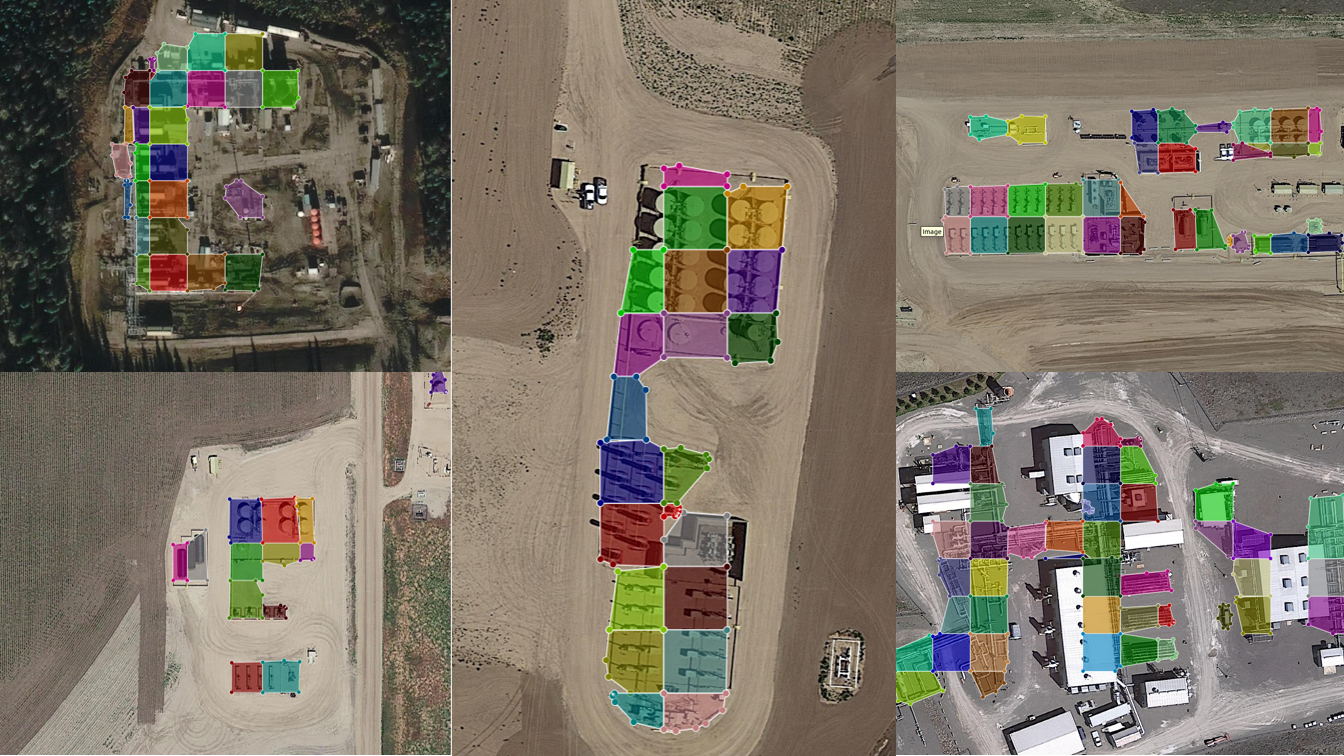}
    \caption{\textbf{Qualitative Results. } Examples of high-quality subspace extraction from visually dissimilar satellite images using SmartScan.}
    \label{fig:qual}
\end{figure}

Figure \ref{fig:qual} shows qualitative results of high-quality subspace extraction using SmartScan from significantly visually dissimilar satellite images. This shows the combined effect of few-shot capabilities of our Autonomous Prompting System, and the domain-agnostic zero-shot efficiency of SAM.

\section{Conclusion}\label{Sec:6}
We presented SmartScan, an AI-based framework that automates the extraction of pertinent data-sets necessary for optimal sensor placement design. Thus, the subspaces of interest are extracted from a satellite image of the site with an interactive tool that helps to quickly create facility-specific and problem-dependent constraint
sets (including bounds, perimeter, exclusion zones and other constraints).

SmartScan employs the Segment Anything Model (SAM),
a prompt-based transformer model for zero-shot segmentation
for extracting sub-spaces (regions of interest) from any satellite image without need for explicit training. 
SmartScan is streamlined for producing
high-quality sub-space extraction with high throughput and
minimal human supervision (quality check) with its novel end-to-end design and AI-based prompting mechanism, thus increasing scalability and efficiency of downstream algorithms. Further, the design of SmartScan makes it suited for extracting regions of interest from any ultra high-resolution satellite imagery, making it domain agnostic. SmartScan shows the utility of accurately prompting a pre-trained segmentation model such as SAM, thereby negating the need for a fully-supervised, task-specific segmentation model trained from scratch.
\section{Acknowledgments}
This work would not have been possible without the support of several colleagues at Schlumberger-Doll Research during my internship. 
I would like to thank the Sensing and Emissions team, and in particular, Andrew Speck, Junyi Yuan, Anna Tifft, Jafet Ruiz Santana and Antonio Vieira for their unwavering help.
Lastly, I would like to extend my sincere thanks to Kashif Rashid, my mentor, for his valuable ideas and immense support for the entire duration of this project. 
\bibliographystyle{IEEEtran}
\bibliography{egbib}

\end{document}